\pdfoutput=1

\documentclass[11pt]{article}

\usepackage{acl}

\usepackage{times}
\usepackage{latexsym}

\usepackage[T1]{fontenc}

\usepackage[utf8]{inputenc}

\usepackage{microtype}

\usepackage{graphicx}
\usepackage{enumitem}
\usepackage{url}
\usepackage{booktabs}
\usepackage{siunitx}
\usepackage{etoolbox}
\robustify\bfseries
\title{DialogueScript: Using Dialogue Agents to Produce a Script\footnotemark}

\author{
  Patrícia Schmidtová, 
  Dávid Javorský,
  Christián Mikláš, 
  Tomáš Musil, 
  Rudolf Rosa, 
  Ondřej Dušek \\
  Charles University, Faculty of Mathematics and Physics,
Prague, Czechia \\
  }

\date{}

\begin{document}
\maketitle
\renewcommand{\thefootnote}{\fnsymbol{footnote}}
\setcounter{footnote}{1}
\footnotetext{Non-Archival WNU submission.}
\renewcommand{\thefootnote}{\arabic{footnote}}
\setcounter{footnote}{0}

\begin{abstract}
We present a novel approach to generating scripts by using agents with different personality types. To manage character interaction in the script, we employ simulated dramatic networks. Automatic and human evaluation on multiple criteria shows that our approach outperforms a vanilla-\emph{GPT2}-based baseline. We further introduce a new metric to evaluate dialogue consistency based on natural language inference and demonstrate its validity.
\end{abstract}


\section{Introduction}
The last couple of years have seen some promising advancements in the area of open-ended story generation \citep{fan_hierarchical_2018, clark_neural_2018, ammanabrolu_guided_2019}, notably with the use of large pretrained generative neural language models such as GPT-2 \cite{radford2019language,see-etal-2019-massively}. 
However, these works mostly focus on producing very short stories, such as those in ROCstories \citep{rocstories}. While there have been attempts at generating full-length theatrical works involving longer dialogue scripts, they use human-in-the-loop approaches, such as post-editing \cite{colton2016beyond,helper2018} or human choice between alternatives during the generation process \cite{theaitre}. 
Longer texts fully generated by language models \cite{sharp_sunspring_2016} often show as inconsistent and/or dull.


In this work, we explore a novel approach to generating longer scripted dialogues, such as theatre or movie scripts, inspired by works in personalizing dialogue agents \citep{zhang-etal-2018-personalizing, mazare-etal-2018-training}. Instead of handcrafting specific personas such as these previous works,  
we propose to cluster personalities using an approximation by the prevailing sentiment in the respective characters' utterances. We use these clusters to train three distinct models, which then act as a positive, neutral and a negative character. Since there are more than two characters, we need a non-trivial dialogue management system do decide the order of characters in the dialogue. We design a novel approach based on simulating dramatic networks \citep[DN;][]{SimulatingDramaticNetworks}.
We compare our overall script generation approach to a baseline based on a vanilla GPT-2 model \cite{radford2019language}.
We use basic automatic metrics for diversity and sentiment, combined with human evaluation on multiple criteria.
Since automatic metrics for evaluating coherence of open-ended text generation are scarce, we present a new automatic metric based on natural language inference \citep[NLI;][]{mnli}.

Our contributions include: (1) DialogueScript -- script generation with distinct language models for different characters, based on character clustering; (2) dialogue management based on DN; (3) NLI-Score -- a novel metric for the evaluation of consistency of the generation outputs; and (4) automatic and human evaluation comparing our DialogueScript/DN approach to a vanilla GPT-2 baseline.
We plan to release our experimental code and models on GitHub.\footnote{Link will be provided in the final version of the paper.}

\section{Script Generation Approach}
\label{sec:model}

\subsection{Character Clustering}
\label{sec:clustering}

Characters in movies usually display a consistent personality within their utterances. However, training models for specific characters would make it difficult to explore various genres or situations due to training data sparsity. To find an acceptable balance between consistency and versatility, we simplify the training and group characters into several disjoint subsets based on their prevailing utterance sentiment.
This selection is realized by a sentiment classifier by \cite{barbieri2020tweeteval},\footnote{\url{https://huggingface.co/cardiffnlp/twitter-roberta-base-sentiment}} which is based on a pre-trained \textit{RoBERTa-base} model \citep{liu2019roberta}, further trained on masked language modeling on on 58M tweets and finetuned on tweet sentiment classification.
The model classifies the input into three groups, labeling it as positive, neutral or negative. Because the input length of the model is limited, processing all utterances of a character glued together would cause an undesirable input truncation. To address this issue, we label each dialogue turn individually and the overall character cluster assignment is computed as the prevailing sentiment over all their utterances.

\subsection{Data Preprocessing for Language Models}
\label{sec:preprocessing}

Before the individual character language models are trained, the dataset needs to be pre-processed. Since the characters are identified by their membership in a cluster instead of their names, the name of each character is replaced by the label \emph{focus} or \emph{other}. The former denotes that the type of this character matches the sentiment of the model, e.g.\ when training a positive model, a positive character is labeled as the focus. The latter is used for marking characters that are not salient for current learning, e.g.\ the label of the negative and neutral characters in training data for the positive model. Because multiple characters within the same cluster may occur in one dialogue, several instances of every dialogue with different focus/other labels are included in the training data.

\subsection{Simulating Dramatic Networks}
\label{sec:moretti}

To orchestrate the script generation between the separate character language models generating individual utterances, we design a new approach based on DN \cite{SimulatingDramaticNetworks}.
We consider the script/dialogue to consist of one or more exchanges (one character starting and others replying) and each line to be addressed to one specific character (i.e., character A utters a line addressed to character B).
The dialogue flow is determined by interpretable parameters of characters and their relations. 
There are 3 main parameters per character: 
\begin{itemize}[leftmargin=10pt,parsep=0pt,itemsep=1pt,topsep=1pt]
\item \emph{centrality} -- the probability of addressing another character (starting an exchange), 
\item \emph{loyalty} -- probability distribution over potential addressees,
\item \emph{reciprocity} -- probability of replying to an address.
\end{itemize}
All parameters are updated throughout the script generation.
Unlike \citet{SimulatingDramaticNetworks}, we do not estimate model parameters from existing play scripts. Instead, we set initial model parameters empirically based on a few test trials, and we use the DN model to manage generation of new scripts.\footnote{Moreover, while \citet{SimulatingDramaticNetworks}'s approach considers multiple scenes, we only assume a single scene/dialogue for simplicity.} 

While all characters initially have the same centrality (i.e., the probability of starting the dialogue, set at 1), centrality increases with every line spoken by the given character.
At the end of the script, each character's centrality reflects their significance for the generated script.

The loyalty parameter works similarly -- if character A addresses character B at a given point in the script, their probability of addressing B in the future increases (at the expense of other characters).
At the end of the script, the loyalty probability distribution reflects relationships a certain character had with all other characters. 
We set the loyalty probability distribution uniformly.

The reciprocity parameter determines if B responds to A after being addressed. 
To present a realistic length of exchanges between two characters in the script, reciprocity starts at 95\% and decays by a third after each line uttered. The initial value and the decay rate are defined separately for each character. They determine the length of exchanges between two certain characters and reflect characters' talkativeness. Reciprocity resets after the end of a given exchange (when B decides not to respond to A). When an exchange ends, the next character to speak is chosen by centrality.

The probability of the dialogue ending after each line is independent of characters' relations; it is fixed at 20\% throughout the generation.

%
%
%
%

\section{Evaluation Metrics}
\label{sec:eval}

Since standard reference-based language generation metrics are not applicable to our free-form long-text generation scenario,
we combine basic corpus-based statistics showing diversity with evaluation of personality consistency via sentiment classification, coupled with human evaluation based on multiple criteria. We also propose a new automatic metric targeted at consistency.

\subsection{Automatic Metrics}
\label{sec:autom}

\paragraph{Diversity} We evaluate several automatic metrics aimed at text diversity  \cite{van_miltenburg_measuring_2018}.
This includes the perplexity, the total number of words generated, as well as the number of distinct words (1-grams) and bigrams. All diversity metrics are measured as average over generated dialogues.

\paragraph{Personality consistency} To show that our models can generate consistent utterances based on the target character types, we measure sentiment of utterances in the generated dialogues, similarly to the training data clustering approach from Section~\ref{sec:clustering}.

\subsection{Human Evaluation}
\label{sec:manual}

We design two manual evaluation procedures, both to be carried out on the same text samples to reduce annotator mental load:

\paragraph{Relative ranking}
The annotators are asked to order dialogues generated by different systems from best to worst, according to their own subjective judgement, with no further instructions. This ranking gives us an overall system comparison. 

\paragraph{Absolute scoring}

The annotators are asked to rate the generated dialogues in terms of the following properties on a 5-point Likert scale:
\begin{itemize}[leftmargin=10pt,parsep=0pt,itemsep=1pt,topsep=1pt]
    \item \emph{Coherence}: Is the text coherent?
    \item \emph{Consistency}: Are the characters self-consistent? 
    \item \emph{Originality}: Is the text original and interesting?
    \item \emph{Overall impression}: Did you enjoy reading this text?
\end{itemize}


\subsection{NLI-Score: A Consistency Metric}
\label{sec:nli-score}

Inspired by previous approaches using NLI to evaluate texts for other NLG tasks \cite{dziri_evaluating_2019,maynez_faithfulness_2020}, we develop NLI-Score, a new metric for dialogue consistency.

In general, NLI determines whether a given sentence is entailed in, neutral to, or in contradiction with a context \cite{bowman_large_2015}.
Unlike previous works, we aim at the \emph{neutral} relation in NLI-Score, which indicates newly added information, but no inconsistencies.
The contradiction relation indicates inconsistencies and the entailment relation is mostly indicative of repetition, both of which are unwanted in creative text generation.
We use the \textit{RoBERTa-large-mnli} model by \cite{liu2019roberta}\footnote{\url{https://huggingface.co/roberta-large-mnli}} to compute probabilities of the different NLI classes, then take the probability of the neutral category as the basis our NLI-Score.
To make the metric robust to varied length, we propose to measure the average neutrality per added sentence.
The second sentence is compared with the first, the third with the first two, and so on.\footnote{The context is truncated from the start if its length exceeds the NLI model's maximum input length.}

%
%
%
%

\section{Experiments}

\begin{table*}[t]
\centering
\small
\begin{tabular}{lccccc}
\toprule
\textbf{Model} & \textbf{Perplexity} & \textbf{1-gram Vocab} & \textbf{2-gram Vocab} & \textbf{Words} & \textbf{NLI-Score}\\
\midrule
Baseline &1.86 & \phantom{0}59.25 & \phantom{0}88.00 & 104.00 & 0.40 \\
DialogueScript & \bf 2.48 & \bf 241.90 & \bf 428.06 & \bf 489.76 & \bf 0.47\\
DialogueScript + DN & 2.13 & 183.57 & 308.13 & 359.61 & 0.46\\
\bottomrule
\end{tabular}
\caption{Automatic metric results: generated script diversity (average perplexity, unigram and bigram vocabulary size, number of words) and consistency in terms of NLI-Score (see Section~\ref{sec:nli-score}).}\label{tab:automatic}
\end{table*}

\begin{table}[t]
    \centering
    \small
    \begin{tabular}{lrrr}
    \toprule
    & & \textbf{Sentiment} & \\
    \textbf{Character} &\textbf{Positive} & \textbf{Neutral} & \textbf{Negative}  \\
    \midrule
     Positive & 35 & 64 & 1 \\
     Neutral & 1 & 14 & 1 \\
     Negative & 4 & 56 & 40 \\
     \bottomrule
    \end{tabular}
    \caption{Sentiment of the generated utterances, depending on the target sentiment for a given character.}
    \label{tab:sentiment}
\end{table}

\begin{table}[t]
    \small
    \centering
    \begin{tabular}{lccc}
    \toprule
    \textbf{Model} &\textbf{1st} &\textbf{2nd} & \textbf{3rd} \\
    \midrule
    Baseline & 4 & 2 & 6 \\
    DialogueScript & 2 & 5 & 5 \\
    DialogueScript + DN & 6 & 5 & 1 \\
    \bottomrule
    \end{tabular}
    \caption{Results of relative ranking of model outputs.}
    \label{tab:ranking}
\end{table}

\begin{table}[t]
    \small
    \centering
    \begin{tabular}{lcccc}
    \toprule
    \textbf{Model} &\textbf{Coh} &\textbf{Con} & \textbf{Orig} & \textbf{Overall} \\
    \midrule
    Baseline & 2.3 & 2.7 & 2.5 & 2.5 \\
    DialogueScript & \textbf{3.3} & 3.2 & 3.8 & 3.3 \\
    DialogueScript + DN & 3.0 & \textbf{3.3} & \textbf{4.7} & \textbf{3.8}\\
    \bottomrule
    \end{tabular}
    \caption{Average absolute human rating scores -- \underline{Coh}erence, \underline{Con}sistency, \underline{Orig}inality and \underline{Overall} impression, on a 5-point Likert scale.}
    \label{tab:absolute}
\end{table}

\subsection{DialogueScript Training}
\label{sec:training}

In DialogueScript, characters are represented by three separate language models trained by fine-tuning the \textit{GPT2-small} model \citep{radford2019language}, given the respective clustered data (positive, neutral, or negative) as described in Sections~\ref{sec:clustering} and~\ref{sec:preprocessing}. The training uses an adaptive learning rate optimizer ($\alpha = 3 \times 10^{-5}$, $\epsilon = 1 \times 10^{-8}$) \citep{kingma2015adam} and a linear scheduler with warmup of 1,000 steps over five epochs. 

To finetune the models, we use a dataset consisting of movie scripts (1,276 movies) from ScriptBase \cite{scriptbase} and TV show scripts (786 episodes) scraped from fan-sourced collections,   IMSDb\footnote{\url{https://imsdb.com/}} and Forever Dreaming.\footnote{\url{https://transcripts.foreverdreaming.org/}}

\subsection{Compared Model Variants}

We evaluate 3 model variants: (1) a base \emph{DialogueScript} model with random order of characters, (2) an extended \emph{DialogueScript + DN} (based on the DN orchestration described Section~\ref{sec:moretti}), and (3) a \emph{Baseline} based on vanilla \emph{GPT2-medium} for comparison.
Every generated dialogue includes three characters (each supposedly corresponding to one character type, i.e.~positive, neutral and negative).

Both DialogueScript setups receive no textual initialization and generate scripts from scratch.
This is not possible with the baseline, which requires a prompt to generate a script-like text.\footnote{In our experiments, the unprompted \textit{GPT-2} model generated HTML code.}
Therefore, we use minimal prompts (a short 1-sentence setting description + single-utterance greeting from all three characters) to start the baseline model generation. These prompts are not included in the evaluation.

Note that the DialogueScript and DialogueScript + DN systems differ only in the order of the characters' utterances and the length of scenes.
The dialogue management does not influence the content of the utterances themselves in any way, their content is generated using the same sentiment-based models (see Section~\ref{sec:training}).

\subsection{Results}

\paragraph{Automatic metrics} For automatic evaluation, we use 50 scripts generated by our systems and 10 scripts by the \textit{GPT2-medium} baseline.
Table~\ref{tab:automatic} shows that both DialogueScript setups produce more diverse scripts than the baseline.
Table~\ref{tab:sentiment} then demonstrates the inclination of DialogueScript model outputs to their target sentiment, with the exception of a prominent neutral sentiment. This is natural, because we cannot expect the characters to avoid common phrases with a neutral sentiment. 

\paragraph{Human evaluation}
We use 12 short excerpts from scripts generated by each model for all of the manual evaluation tasks. 
The annotators are shown 5-10 lines\footnote{The amount of text was similar for all evaluated dialogues as the number of lines was balanced by their length.} at a time. Each annotation is performed by 3 judges.

Table~\ref{tab:ranking} with relative ranking results shows that DialogueScript + DN was most frequently the best option and least frequently the worst one. 
As we can see in Table~\ref{tab:absolute}, both our systems beat the baseline in all of the absolute scoring criteria. The DialogueScript + DN setup scores better than base DialogueScript with random character ordering on all criteria except Coherence. Since both DialogueScript setups use the same models, we believe that the DN orchestration made a difference in making the character interaction more organic.

\paragraph{NLI-Score} 
We evaluated our new metric by comparing it to human evaluation of consistency. The scores have a Pearson correlation of 0.50, showing that NLI-Score does provide some consistency information.
When we apply NLI-Score for automatic evaluation of the compared setups (see Table~\ref{tab:automatic}), we can see that NLI-Score is similar for both DialogueScript approaches and in both cases higher than the baseline, showing that our generated texts contain less detectable contradictions and repetitions than the baseline.

\subsection{Discussion}
 
 While metrics such as perplexity can characterize an NLG output, they are not enough to decide on the overall output quality. However, we can use these characteristics to make an observation that our systems tend to be more verbose than the baseline approach. We hypothesize that this might have played a role in the human evaluation, especially in the ranking task where the baseline texts appeared sleeker and therefore easier to read.

\section{Conclusion}


We approached script generation by simulating the interaction of characters. We prepared training data for three different personalities/sentiments (positive, neutral and negative) by clustering average sentiment values of characters in movies and TV shows. We trained the corresponding models and combined them by simulating dramatic networks. We proposed a new metric, the NLI-Score, to automatically evaluate the consistency of the generated text. Based on both automatic metrics and human evaluation, our approach outperforms the baseline. Our NLI-Score metric shows as indicative of overall output consistency.




\bibliographystyle{acl_natbib}
\bibliography{emnlp2020}

\end{document}